\documentclass[fleqn,10pt]{wlscirep}
\usepackage[utf8]{inputenc}
\usepackage[T1]{fontenc}

\title{RAMIE: Retrieval-Augmented Multi-task Information Extraction with Large Language Models on Dietary Supplements}

\author[1]{Zaifu Zhan, MEng}
\author[2]{Shuang Zhou, PhD}
\author[2]{Mingchen Li, MS}
\author[2,*]{Rui Zhang, PhD}
\affil[1]{Department of Electrical and Computer Engineering, University of Minnesota, Minneapolis, MN, USA}
\affil[2]{Division of Computational Health Sciences, Department of Surgery, University of Minnesota, Minneapolis, MN, USA}

\affil[*]{Corresponding author: Rui Zhang, PhD, Division of Computational Health Sciences, Department of Surgery, University of Minnesota, 11-132 Phillips-Wangensteen Building, 516 Delaware St SE, Minneapolis, MN 55455, United States (\href{zhan1386@umn.edu}{zhan1386@umn.edu})}

\begin{abstract}
\textbf{Objective:} We aimed to develop an advanced multi-task large language model (LLM) framework to extract multiple types of information about dietary supplements (DS) from clinical records.

\textbf{Methods:} We used four core DS information extraction tasks—namely, named entity recognition (NER: 2,949 clinical sentences), relation extraction (RE: 4,892 sentences), triple extraction (TE: 2,949 sentences), and usage classification (UC: 2,460 sentences) as our multitasks. We introduced a novel Retrieval-Augmented Multi-task Information Extraction (RAMIE) Framework, including: 1) employed instruction fine-tuning techniques with task-specific prompts, 2) trained LLMs for multiple tasks with improved storage efficiency and lower training costs, and 3) incorporated retrieval augmentation generation (RAG) techniques by retrieving similar examples from the training set. We compared RAMIE’s performance to LLMs with instruction fine-tuning alone and conducted an ablation study to assess the contributions of multi-task learning and RAG to improved multitasking performance.

\textbf{Results:} With the aid of the RAMIE framework, Llama2-13B achieved an F1 score of 87.39 (3.51\% improvement) on the NER task and demonstrated outstanding performance on the RE task with an F1 score of 93.74 (1.15\% improvement). For the TE task, Llama2-7B scored 79.45 (14.26\% improvement), and MedAlpaca-7B achieved the highest F1 score of 93.45 (0.94\% improvement) on the UC task. The ablation study revealed that while MTL increased efficiency with a slight trade-off in performance, RAG significantly boosted overall accuracy.

\textbf{Conclusion:} This study presents a novel RAMIE framework that demonstrates substantial improvements in multi-task information extraction for DS-related data from clinical records. Our framework can potentially be applied to other domains.
\end{abstract}

\begin{document}

\flushbottom
\maketitle
%
%
 \thispagestyle{empty}

\section{Introduction}
Dietary supplements (DSs) play a pivotal role in promoting health and wellness by providing essential nutrients that may be lacking in regular diets. According to the 2023 CRN consumer survey~\cite{crn2023survey}, 74\% of U.S. adults use supplements, with 55\% being regular users, and 92\% of these users agree that DSs are essential for health. Despite their widespread acceptance, concerns persist regarding the quality~\cite{fu2009quality}, effectiveness~\cite{dodge2011influence}, and safety\cite{dodge2011influence,petroczi2011mission} of DSs due to their classification as food rather than medicine, exempting them from FDA approval~\cite{fda2023facts}. This lack of regulation leads to several issues, including insufficient transparency regarding ingredient identities~\cite{tucker2018unapproved}, a shortage of rigorously designed clinical trials~\cite{d2018effect}, and limited laboratory studies clarifying their mechanisms~\cite{van2019limitations}, which could result in adverse events (AEs)~\cite{palmer2003adverse}, some of which may be severe or even fatal~\cite{timbo2018dietary}. 

Clinical records contain extensive DS information and their AEs~\cite{fan2017detecting,zhang2015evaluating,fan2018using}, offering valuable insights for public health, medical research, and regulation. 
However, this information is often embedded in unstructured text within electronic health records~\cite{Shickel2018DeepEHR,cowie2017electronic,jha2009use}, requiring advanced information extraction methods to comprehensively and accurately identify DS entities, related events, and their interrelations~\cite{fan2021deep}.
Several tasks in extracting DS information include named entity recognition (NER)  for identifying specific DS names~\cite{lee2021multiple}, relations extraction (RE) for discerning associations between DSs and AEs versus indications or no relations~\cite{zhou2024leap,li2024benchmarking}, triple extraction (TE) or end-to-end NER with RE, and usage classification (UC) for capturing DS usage status (such as start, continuation, or discontinuation)~\cite{fan2018using}.
Moreover, the variability in clinical language~\cite{holper2020ambiguous,grossman2021deep}, including misspellings~\cite{fan2019using}, abbreviations~\cite{rizvi2020idisk}, and ambiguous terms, adds to the complexity of automated information extraction. 
Given these challenges, there is a pressing need for advanced methods that can handle the complexity and diversity of DS-related information embedded in clinical records. 

To solve these challenges, there has been some prior research on using natural language processing (NLP) techniques to analyze DSs in text~\cite{dang2020ensemble,singh2022conversational,schutte2021discovering}. For example, Fan et al. used Bi-LSTM and BERT models to recognize DS entities, extract relations between DSs and AEs from clinical text~\cite{fan2021deep}, and predict DSs use status~\cite{fan2018using}. 
Zhou et al.~\cite{zhou2021identification} trained BERT and its variants to identify DS use status from clinical notes among patients with Alzheimer’s disease and related dementias. 
However, due to the analytical limitations of smaller models such as CNNs and BERT, these approaches often fall short in handling unseen text or complex clinical text, particularly when required to process multiple entity types and intricate relationships essential for comprehensive DS analysis.
Recent advancements in LLMs such as GPT~\cite{radford2018improving, radford2019language, brown2020language} and the Llama series~\cite{touvron2023llama} have revolutionized the artificial intelligence domain~\cite{bent2023large} and many efforts have reported the effectiveness of applying LLMs for health records~\cite{zhou2024large} and information extraction tasks~\cite{goel2023llms,monajatipoor2024llms,zhou2024leap,chen2024large,sun2024consistency,zhang2024pushing}. 
However, LLMs' application (and with advanced techniques such as RAG and instruction Fine-tuning) to DS-related information extraction remains under-explored. 
Although Zhou et al.~\cite{zhou2024complementary} employed GPT-3.5-turbo to extract DS-related entities as part of broader therapies, LLM's ability to extract DS entities is still unknown.
To fill the gap, we investigate the potential of LLMs for comprehensive information extraction tasks in the DS domain as the first benchmark work.

In this work, considering that the efficiency of information extraction is essential in real-world scenarios~\cite{zhao2021unified, zheng2017joint, qin2022enhancing}, we propose the RAMIE framework (\textbf{R}etrieval-\textbf{A}ugmented \textbf{M}ulti-task \textbf{I}nformation \textbf{E}xtraction Framework) for Dietary Supplements via Multi-Task Learning (MTL) ~\cite{chen2024multi, ruder2017overview} and RAG to effectively handle multiple information extraction tasks simultaneously.
On the one hand, MTL~\cite{zhao2021unified, zheng2017joint, qin2022enhancing} offers significant efficiency, reducing not only training costs but also storage requirements~\cite{kang2011learning, yin2024llm}, laying the foundation for deploying healthcare applications on smaller devices in the future. 
For example, Mulyar et al.~\cite{mulyar2021mt} sacrificed little performance to achieve multi-head BERT to tackle multiple tasks in one model.
On the other hand, RAG~\cite{lewis2020retrieval} offers a promising avenue to enhance LLMs for DSs. By incorporating retrieval mechanisms, RAG allows models to access additional relevant examples during generation~\cite{li2024benchmarking, wu2024stark}. 
For instance, RAG has been successfully applied to improve question-answering systems ~\cite{hou2024enhancing}.

Our contributions are summarized as follows,
\begin{itemize}
    \item To the best of our knowledge, it is the first work to explore the potential of LLMs for information extraction multitasks in the DS domain, including NER, RE, TE, UC tasks.
    \item We proposed the RAMIE (\textbf{R}etrieval-\textbf{A}ugmented \textbf{M}ulti-task \textbf{I}nformation \textbf{E}xtraction) framework which demonstrated a high performance via RAG and instruction fine-tuning, and achieve efficiency via MTL.
    \item We conducted comprehensive experiments on 8 state-of-the-art LLMs through 1) single-task instruction fine-tuning, 2) our RAMIE framework, and 3) multi-task instruction fine-tuning.
\end{itemize}

\begin{figure}[t]
    \centering
\includegraphics[width=0.9\linewidth]{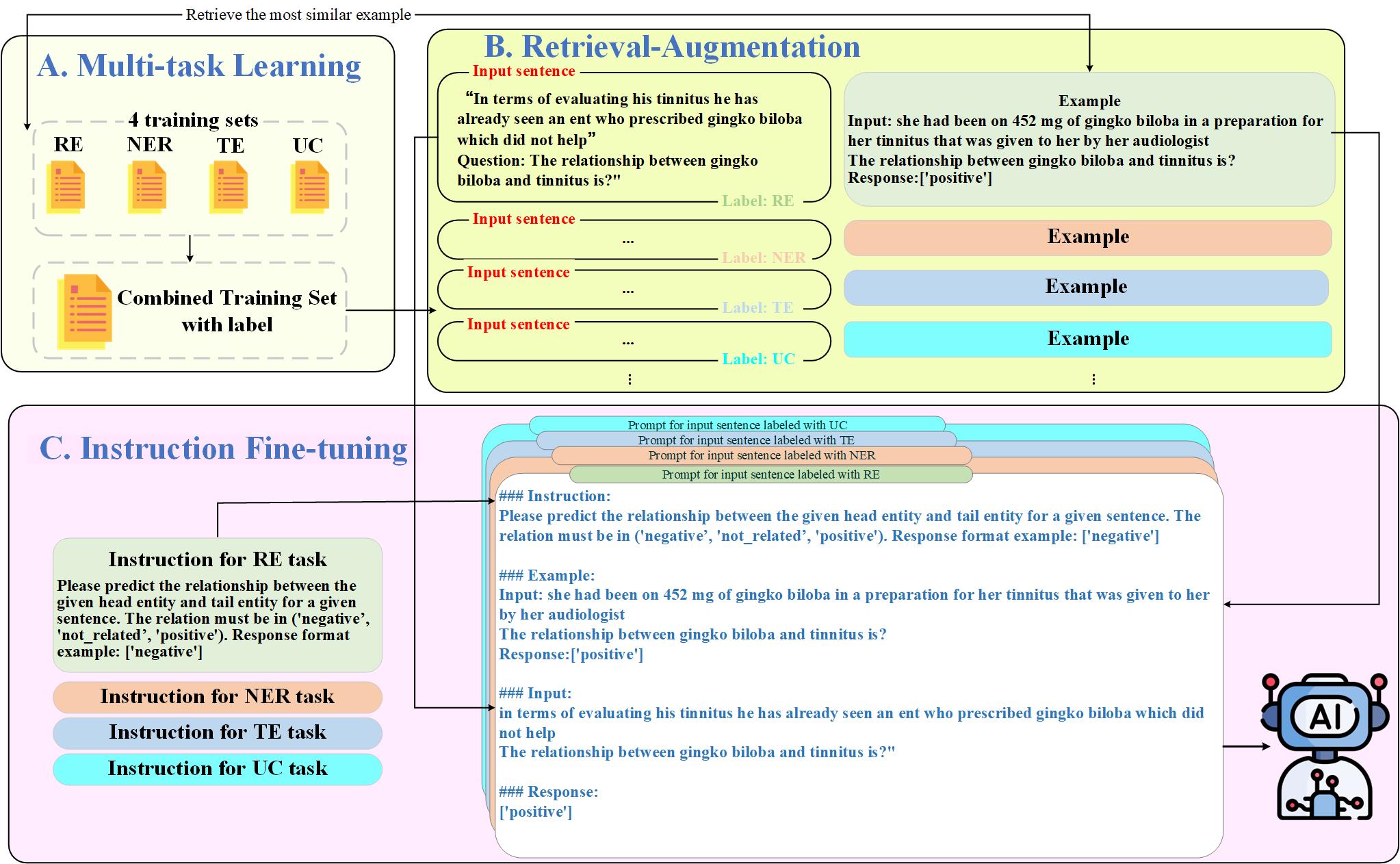}
    \caption{The retrieval-augmented multi-task information extraction framework(RAMIE).}
    \label{fig:framework}
\end{figure}


\begin{table}[h]
    \centering
    \begin{tabular}{cp{15cm}}
        \hline
        \textbf{Task} & \multicolumn{1}{c}{\textbf{Statistics / Annotations / Examples}} \\
        \hline
        NER & 
        Train / Dev / Test size: 2365 / 292 / 292
        \newline DS entities: \textit{folic\_acid, milk\_thistle, ginger, chamomile, garlic, black\_cohosh, ginkgo, lavender, melatonin, cranberry, glucosamine, dandelion, saw\_palmetto, green\_tea.} 
        \newline All adverse events were labeled as an \textit{event}
        \newline Examples:
        \newline - \textit{ginger} and \textit{ginkgo biloba:} ginger and ginkgo biloba are mild anticoagulants so there is an increased risk of bleeding, especially with full-dose aspirin
        \newline - \textit{green tea}: She did not state that she started drinking a lot of green tea before was noted her blood pressure was going up
        \newline - \textit{lavender:} pt expressed anxiety but comforted with support breathing and lavender
        \newline - \textit{ginseng:} constipation-use of ginseng tea-seems to be better with this
        \\
        \hline
        RE & 
        Train / Dev / Test size: 3964 / 464 / 464
        \newline Relations: \textit{positive, negative, not\_related}, indicating the relation between the head DS entity and the corresponding adverse events. 
        \newline Examples:
        \newline - \textit{positive:} ginkgo biloba and melatonin have both been studied for use in treating tinnitus . The relationship between melatonin and tinnitus is?
        \newline - \textit{negative:} she has tried melatonin but it is increased the morning dizziness . The relationship between melatonin and dizziness is?
        \newline - \textit{not\_related:} will add maalox for indegestion ,fiber to decrease constipation,and lavender aromatherapy for pain /anxiety relief .The relationship between fiber and pain is?
        \\
        \hline
        TE & 
        Train / Dev / Test size: 2365 / 292 / 292
        \newline Same entities as NER task (head entities are DS entities and tail entities are the corresponding adverse events), relation types are same as RE task.
        \newline Examples:
        \newline - \textit{ginseng tea|positive|constipation:} constipation-use of ginseng tea-seems to be better with this
        \newline - \textit{estroven|positive|night sweat:} She's taking estrogen for night sweats' it helps a little.
        \newline - \textit{folic acid|negative|rash:} daughter calling back to inform this office that they accidentally purchased folic acid 400mg and have been giving 1 po to pt since 09-12-08. daughter is asking if her recent skin rash could be due to this.
        \\
        \hline
        UC & 
        Train / Dev / Test size:  2000 / 230 / 230
        \newline Use status: \textit{continue, discontinue, uncertain, start}, describing the status of DS for patients.
        \newline Examples:
        \newline - \textit{continue:} continue Vitamin E selenium discharge
        \newline - \textit{discontinue:} note stop b6 b12 folic acid today
        \newline - \textit{uncertain:} suggest take co mg daily vitamin d3 unit daily
        \newline - \textit{start:} currently prescribe levo thyroxine mcg daily melatonin
        \\
        \hline
    \end{tabular}
    \caption{Overview of annotation details and the statistics of datasets.}
\label{tab:data}
\end{table}



\section{Methods}

\subsection{Overview of methods} 
In this work, we propose a comprehensive framework for performing multi-task information extraction on clinical narratives related to DSs. Our framework, RAMIE (Retrieval-Augmented Multi-task Information Extraction), is designed to conduct four tasks including NER, RE, TE, and UC within a single model. 
RAMIE combines MTL, RAG, and instruction fine-tuning to optimize extraction accuracy, model efficiency, and scalability across tasks. 
The framework has the potential to be applied to other domains.
By using instruction-tuning and integrating retrievers to enhance context relevancy, RAMIE demonstrates adaptability to the complexities of clinical text analysis.
In subsequent sections, we describe the tasks, datasets, architectural innovations that underpin RAMIE's capabilities, and experimental setup.

\subsection{Tasks and Datasets} 
In this work, we address four core information extraction tasks: NER~\cite{lee2021multiple}, RE~\cite{zhou2024leap}, TE~\cite{li2024benchmarking}, and UC~\cite{fan2018using}. These tasks are foundational to NLP for extracting structured data from unstructured clinical text. Each task serves a specific role in transforming clinical text into a format that can support downstream applications, such as clinical decision-making or research on DSs. \begin{enumerate} \item \textbf{NER}: This task involves identifying and categorizing DS entities and AEs in the text. The model detects mentions of supplements and associated events in clinical notes, marking the entities with predefined categories. For instance, in a sentence like "The patient reported taking cranberry juice for a urinary tract infection", the model would tag "cranberry juice" as a dietary supplement and "urinary tract infection" as an adverse event.

\item \textbf{RE}: Once entities are identified, this task determines the relationships between them. In the context of DS-related data, RE identifies whether a supplement is positively or negatively associated with an event, or if there is no direct relationship. This task enables the identification of cause-effect relationships or therapeutic uses of DSs. For example, in "The patient experienced nausea after taking ginseng", RE would identify a negative relationship between "ginseng" and "nausea."

\item \textbf{TE}: TE is an end-to-end RE task by structuring information into subject-predicate-object triples directly from the inputs. This task identified the named entities and then determined the relationships between identified entities into triples, which can be used for further analysis, such as creating knowledge graphs. In a sentence like "Cranberry is used to prevent urinary tract infections", TE would extract the triple (Cranberry <has\_indication> urinary tract infections).

\item \textbf{UC}: UC focuses on classifying the use status of dietary supplements described in clinical notes. The model identifies whether a supplement is "Started", "Continued", "Discontinued", or its status is "Uncertain" within a given sentence. For example, in "The patient stopped taking fish oil due to side effects", the model would classify the use status as "Discontinued." \end{enumerate}

The datasets for these 4 tasks were derived from the Clinical Data Repository (CDR) at the University of Minnesota. The CDR contains over 180 million clinical notes from more than 5 million patients~\cite{fan2021deep}, providing a rich source of information for DS-related tasks. Each task is supported by a specific dataset, which was manually annotated by domain experts~\cite{fan2021deep}. 
The dataset was split into training, development, and test sets with a ratio of 8:1:1.
The data statistics, annotation details, and examples for each task are summarized in Table~\ref{tab:data}.

\subsection{RAMIE framework}
The motivation for to proposed RAMIE framework is to maintain the high performance of LLMs while achieving efficiency.
As depicted in Fig. \ref{fig:framework}, the RAMIE framework integrates MTL, RAG, and instruction fine-tuning, leveraging their complementary strengths to enhance task performance. MTL enables LLMs to handle multiple tasks within a single LLM. MTL brings in not only efficiency but also complexity, thus mitigating performance.
The challenge is LLMs should learn to distinguish task types, memorize the response format, and also learn the corresponding task-solving abilities.
Instruction fine-tuning and RAG enhance the performance by providing various instructions for each task and providing task-relevant examples dynamically, respectively.
This combination offers high-level guidance, similar examples, and response templates to enhance understanding and reduce ambiguity in predictions.

To lay a solid groundwork, we evaluate 8 state-of-the-art LLMs in our framework. LLMs include Mistral-7B~\cite{jiang2023mistral}, Llama-2-7B~\cite{touvron2023llama}, Llama-2-13B~\cite{touvron2023llama}, Llama-3-8B~\cite{metaLlama2024} which were pre-trained for general downstream tasks and BioMistral-7B~\cite{labrak2024biomistral}, PMC-Llama-13B~\cite{wu2024pmc}, MedAlpaca-7B~\cite{han2023medalpaca} and MedAlpaca-13B~\cite{han2023medalpaca} which were pre-trained for biomedical domain. 

\subsubsection*{A. Multi-task Learning} 
The MTL in the RAMIE framework offers a practical and scalable solution for deploying LLMs to address diverse tasks, especially in resource-constrained environments. As shown in Fig. \ref{fig:framework}, the RAMIE framework consolidates task-specific datasets for NER, RE, TE, and UC, each containing thousands of labeled sentences, into a unified training set. The task-specific labels serve dual purposes: they not only guide the training process but also play a crucial role in the retrieval and prompt construction phases, where retrieved examples are augmented with task-specific instructions.

The unified dataset is used in a multi-task instruction fine-tuning process, enabling the LLM to be trained simultaneously on sentences from all tasks. During training, the model learns to differentiate between distinct task instructions encoded in the dataset and generates accurate task-specific outputs. By employing MTL approach, RAMIE allows a single LLM to acquire the capability to solve multiple tasks with just one fine-tuning operation. 
This significantly reduces storage redundancy, as only one model needs to be maintained, and minimizes computational resources required for training and deployment. Furthermore, the unified structure and shared learning strategy enhance the adaptability and efficiency of the model across a broad spectrum of information extraction tasks.

\subsubsection*{B. Retrieval-Augmented Generation} 
In the RAMIE framework, as illustrated in Fig. \ref{fig:framework}, we incorporate RAG~\cite{lewis2020retrieval,li2024benchmarking,wu2024stark} to further enhance the model's performance across tasks. During both the training and testing phases, the framework employs retrievers to identify the most relevant sentence-response pair from the corresponding training set based on the cosine similarity of sentence embeddings, determined by the input sentence and its associated task label. To ensure alignment between the retrieved examples and the input format, the retrievers are constrained to search within the training set specific to the input's labeled task.

The retrieval process involves concatenating the input sentence and its corresponding response to form a single representation, which is then compared to the input sentence embedding using cosine similarity. During training, the retrievers often identify pairs where the sentence is identical to the input sentence due to the high embedding similarity. However, such instances may cause the model to rely on copying the response directly from the example, rather than developing analytical reasoning. To mitigate this issue, the retrievers are restricted from selecting the input sentence itself during training and are instead configured to retrieve the most semantically similar example from the remaining dataset. Conversely, during testing, the retrievers are permitted to select the most relevant example from the entire training set, as there is no overlap between the input and retrieved examples.

The retrieved examples are incorporated into a dynamically constructed prompt, which is subsequently used for instruction fine-tuning within the LLM. To optimize retrieval quality, we utilize three state-of-the-art retrievers—MedCPT~\cite{jin2023medcpt}, Contriever~\cite{izacard2021unsupervised}, and BMRetriever~\cite{xu2024bmretriever}. These retrievers are specifically tailored for domain-specific tasks and have demonstrated superior performance in biomedical information extraction, making them well-suited for the diverse and complex tasks addressed in RAMIE.

\subsubsection*{C. Instruction Fine-tuning} 
To optimize the performance of LLMs across multiple tasks, we employed instruction fine-tuning~\cite{longpre2023flan,zhang2023instruction}, a technique that helps models generalize by providing explicit task instructions within the input prompt. Following the LEAP framework~\cite{zhou2024leap}, we designed the template of prompts for each task, as illustrated in Figure~\ref{fig:framework}. The prompt consists of an instruction part, an example part, and the input sentence with its response. The input sentences are from the combined training set, and the examples are retrieved from the retrievers. The instructions are designed to provide instruction for each task as follows, \begin{itemize}
\item For the NER task, the model is instructed to \textit{extract dietary supplements and adverse events from a given sentence and recognize their entity types}. The predefined entity types include \textit{`event', `folic acid', `milk thistle', `ginger', `chamomile', `garlic', `black cohosh', `ginkgo', `lavender', `melatonin', `cranberry', `ginseng', `glucosamine', `dandelion', `saw palmetto', and `green tea'}. The output is expected in the format: \textit{a list of entities with their corresponding types}, for example, \textit{[{entity\_name1: entity\_type1}, {entity\_name2: entity\_type2}, ...]}.

\item In the RE task, the model is prompted to \textit{predict the relationship between a given head entity and tail entity within the provided sentence}. The relation must be selected from the predefined set: \textit{`negative', `not\_related', or `positive'}. The response is expected as \textit{a single-item list}, such as \textit{['negative']}.

\item For the TE task, the model is asked to \textit{extract triples consisting of a head entity, relation, and tail entity from the sentence}. The relation types are the same as in RE, and the entity types are the same as in NER. The output format is \textit{a list of dictionaries}, where each dictionary represents a triple, e.g., \textit{\{'head entity': 'entity\_name', 'relation': 'relation\_type', 'tail entity': 'entity\_name'\}}.

\item In the UC task, the model is required to \textit{predict the dietary supplement usage within the sentence}, selecting from the predefined usage types: \textit{`continue', `discontinue', `uncertain', or `start'}. The expected output is \textit{a list containing the predicted usage type}, such as \textit{['continue']}.
\end{itemize}

\subsection{Experiments} 

We designed three groups of experiments to comprehensively evaluate the proposed RAMIE framework: (1) single-task instruction fine-tuning as a benchmark, (2) the RAMIE framework, and (3) multi-task instruction fine-tuning as an ablation study. The ablation study was designed to provide insights into the individual contributions of MTL and RAG within the RAMIE framework. All experiments were conducted on NVIDIA A100 GPUs with 80GB memory. The training and evaluation batch sizes per device were set to 4, and inference was performed on a sentence-by-sentence basis for efficiency. To fine-tune the model effectively, we employed the LoRA approach~\cite{hu2021lora}, with the rank set to 64, alpha to 32, and a dropout rate of 0.1. The AdamW optimizer was used with a learning rate of 1e-5, and the model was trained for 5000 steps. Evaluations were performed every 1000 steps, and the best-performing model was selected for inference.

For the baseline, we implemented a BERT-based model with separate task-specific heads to handle the four tasks, following the methodology in prior work~\cite{mulyar2021mt}. To evaluate model performance, we adopted Micro Precision, Recall, and F1-score metrics, in line with established studies~\cite{fan2021deep,li2024benchmarking}. A prediction was considered correct only if the entire output exactly matched the ground truth~\cite{fan2021deep,zhou2021identification}. For single-task instruction fine-tuning, models were trained on a single dataset and evaluated on its corresponding test set. In contrast, for multi-task instruction fine-tuning, the models were trained on a blended dataset combining all tasks and evaluated on the test sets of each task individually, both with and without RAG integration. To further evaluate the effectiveness of the retrievers, experiments were conducted using a random retriever as a baseline for comparison.

\begin{table*}[t]
\Huge
\centering
\renewcommand\arraystretch{1.3}
\resizebox{1\textwidth}{!}{%
\begin{tabular} {lccccccccccccc}
\toprule 
\multicolumn{1}{c}  {}& \multicolumn{3}{c}  {Named Entity Recongition} &\multicolumn{3}{c}  {Relation Extraction} &\multicolumn{3}{c}  {Triple Extraction} & \multicolumn{3}{c}  {Usage Classification}\\
\cmidrule(lr){2-4} \cmidrule(lr){5-7} \cmidrule(lr){8-10} \cmidrule(lr){11-13} 
 \multicolumn{1}{c} {Models} & Precision &  Recall & F1 & Precision &  Recall & F1 & Precision &  Recall & F1 & Precision &  Recall & F1 & Avg F1 \\ 
\hline
\textbf{BERT} & 28.08 & 28.08 & 28.08 & 92.88 & 92.88 & 92.88 & 64.71  & 64.57 & 64.64 & 92.61 & 92.61  & \textbf{92.61} & 69.55 \\
\textbf{BioMistral-7B} & 86.01 & 85.89 & 85.95 & 93.09 & 93.09 & 93.09 & 75.00 & 68.48 & 71.59 & 90.79 & 90.79 & 90.79 & 85.36 \\
\textbf{Llama-2-7B} & 85.55 & 81.66 & 83.56 & 83.37 & 83.37 & 83.37 & 72.09 & 64.57 & 68.12 & 89.96 & 89.96 & 89.96 & 81.25\\
\textbf{Llama-2-13B} & 86.40 & 82.34 & 84.32 & 92.66 & 92.66 & 92.66 & 73.01 & 69.66 & 71.29 & 91.70 & 91.70 & 91.70 & 84.99\\
\textbf{Llama-3-8B}  & 86.18 & 84.38 & 85.27 & 93.61 & 93.61 & \textbf{93.61} & 78.29 & 70.39 & 74.13 & 91.01 & 91.01 & 91.01 & 86.01\\
\textbf{MedAlpaca-7B}  & 91.51 & 85.81 & \textbf{88.57} & 89.03 & 89.03 & 89.03 & 81.25 & 70.27 & \textbf{75.36} & 92.57 & 92.57 & 92.57 & \textbf{86.38}\\
\textbf{MedAlpaca-13B}  & 85.01 & 82.90 & 83.94 & 89.42 & 89.42 & 89.42 & 76.05 & 67.78 & 71.68 & 90.83 & 90.83 & 90.83 & 83.97\\
\textbf{Mistral-7B} & 84.01 & 84.83 & 84.42 & 92.66 & 92.66 & 92.66  & 74.88 & 73.33 & 74.09 & 89.08 & 89.08 & 89.08 & 85.06\\
\textbf{PMC-Llama-13B} & 85.37 & 82.90 & 84.11 & 90.50 & 90.50 & 90.50 & 70.46 & 63.82 & 66.97 & 89.08 & 89.08 & 89.08 & 82.67\\
\bottomrule
\end{tabular}
}
\vspace{+2mm}
\caption{Performance of single-task instruction fine-tuning. The highest F1 score for each task and the best average F1 score are presented in bold. } 
\label{tab:stlzt}
\end{table*}

\begin{table*}[!h]
\Huge
\centering
\renewcommand\arraystretch{1.3}
\resizebox{0.99\textwidth}{!}{%
\begin{tabular} {lccccccccccccc}
\toprule 
\multicolumn{1}{c}  {}&\multicolumn{3}{c}  {Named Entity Recongition} &\multicolumn{3}{c}  {Relation Extraction} &\multicolumn{3}{c}  {Triple Extraction} & \multicolumn{3}{c}  {Usage Classification}\\
\cmidrule(lr){2-4} \cmidrule(lr){5-7} \cmidrule(lr){8-10} \cmidrule(lr){11-13} 
\multicolumn{1}{c} {Models} & Precision &  Recall & F1 & Precision &  Recall & F1 & Precision &  Recall & F1 & Precision &  Recall & F1 & Avg F1  \\ 
\hline
\textbf{BioMistral-7B}\\
\quad w/ random example & 85.69 & 79.64 & 82.56 & 90.83 & 90.93 & 90.93 & 64.20 & 60.70 & 62.40 & 82.97 & 82.97 & 82.97 & 79.72\\
\quad w/ MedCPT  & 87.00  & 84.70 & 85.84 & 93.30  & 93.30  & 93.30 & 71.15 & 63.26 & 66.97 & 89.96 & 89.96 & 89.96 & 84.02\\
\quad w/ Contriever & 85.90 & 83.87 & 84.87 & 92.87 & 92.87 & 92.87 & 71.36 & 65.43 & 68.26 & 89.08 & 89.08 & 89.08 & 83.77 \\
\quad w/ BMRetriever & 86.87 & 84.20 & 85.51 & 92.66 & 92.66 & 92.66 & 73.56 & 70.80 & 72.15 & 90.83 & 90.83 & 90.83 & \textbf{85.29}\\
\textbf{Llama2-7B}\\
\quad w/ random example & 78.41 & 81.99 & 80.16 & 82.51 & 82.51 & 82.51 & 62.86 & 64.90 & 63.86 & 69.87 & 69.87 & 69.87 & 74.10\\
\quad w/ MedCPT  & 83.14  & 82.68 & 82.91 & 89.20 & 89.20 & 89.20 & 75.20 & 69.52 & 72.25 & 86.03 & 86.03 & 86.03 & 82.60 \\
\quad w/ Contriever & 83.66 & 83.07 & 83.37 & 90.93 & 90.93 & 90.93 & 67.19 & 65.72 & 66.45 & 90.39 & 90.39 & 90.39 & 82.79  \\
\quad w/ BMRetriever & 84.67 & 84.20 & 84.43 & 88.55 & 88.55 & 88.55 & 70.98 & 90.18 & \textbf{79.45} & 89.08 & 89.08 & 89.08 & \textbf{85.38} \\
\textbf{Llama2-13B}\\
\quad w/ random example & 86.16 & 83.66 & 84.89 & 93.95 & 93.95 & 93.95 & 73.27 & 65.20 & 69.00 & 79.48 & 79.48 & 79.48 & 81.83 \\
\quad w/ MedCPT  & 89.42  & 85.46 & \textbf{87.39} & 92.22 & 92.22 & 92.22 & 75.74 & 70.67 & 73.12 & 86.03 & 86.03 & 86.03 & 84.69 \\
\quad w/ Contriever & 87.35 & 82.61 & 84.92 & 93.52 & 93.52 & 93.52 & 78.99 & 68.27 & 73.24 & 92.14 & 92.14 & 92.14 & \textbf{85.96}  \\
\quad w/ BMRetriever & 87.98 & 85.42 & 86.68 & 93.74 & 93.74 & \textbf{93.74} & 73.13 & 72.45 & 72.79 & 87.28 & 87.28 & 87.28 & 85.12\\
\textbf{MedAlpaca-7B}\\
\quad w/ random example & 77.66 & 84.31 & 80.85 & 91.79 & 91.79 & 91.79 & 64.12 & 68.02 & 66.01 & 87.77 & 87.77 & 87.77 & 81.61 \\
\quad w/ MedCPT  & 85.94 & 84.26 & 85.09 & 88.55 & 88.55 & 88.55 & 72.08 & 69.75 & 70.89 & 90.39 & 90.39 & 90.39 & 83.73 \\
\quad w/ Contriever & 83.33 & 83.45 & 83.39 & 92.01 & 92.01 & 92.01 & 69.45 & 69.15 & 69.30 & 93.45 & 93.45 & \textbf{93.45} & \textbf{84.54}\\
\quad w/ BMRetriever & 82.98 & 86.79 & 84.84 & 92.44 & 92.44 & 92.44 & 67.32 & 71.03 & 69.13 & 89.08 & 89.08 & 89.08 & 83.87 \\

\textbf{MedAlpaca-13B}\\
\quad w/ random example & 78.79 & 81.95 & 80.33 & 87.26 & 87.26 & 87.26 & 56.06 & 56.32 & 56.19 & 80.52 & 81.22 & 80.87 & 76.16 \\
\quad w/ MedCPT  & 86.61 & 84.40 & 85.49 & 87.70 & 88.91 & 88.30 & 64.88 & 66.97 & 65.91 & 86.46 & 86.46 & 86.46 & 81.54\\
\quad w/ Contriever & 83.24 & 83.01 & 83.12 & 90.06 & 92.01 & 91.03 & 69.62 & 64.24 & 66.82 & 92.58 & 92.58 & 92.58 & \textbf{83.39}  \\
\quad w/ BMRetriever & 80.48 & 80.82 & 80.65 & 89.42 & 89.42 & 89.42 & 66.82 & 68.22 & 67.51 & 87.34 & 87.34 & 87.34 & 81.23\\

\textbf{Llama3-8B}\\
\quad w/ random example & 84.91 & 80.96 & 82.89 & 90.93 & 90.93 & 90.93 & 73.24 & 63.62 & 68.09 & 85.15 & 85.15 & 85.15 & 81.77\\
\quad w/ MedCPT  & 88.14 & 83.73 & 85.88 & 92.66 & 92.66 & 92.66 & 71.46 & 76.00 & 73.66 & 88.65 & 88.65 & 88.65 &82.21\\
\quad w/ Contriever & 86.08 & 83.45 & 84.75 & 93.09 & 93.09 & 93.09 & 71.63 & 70.47 & 71.04 & 91.22 & 91.22 & 91.22 & \textbf{85.03}\\
\quad w/ BMRetriever & 88.12 & 84.21 & 86.12 & 92.66 & 92.66 & 92.66 & 72.27 & 73.10 & 72.69 & 88.21 & 88.21 & 88.21 & 84.92\\

\textbf{Mistral-7B}\\
\quad w/ random example & 82.52 & 82.75 & 82.64 & 88.34 & 88.34 & 88.34 & 74.39 & 60.26 & 66.59 & 86.46 & 86.46 & 86.46 & 81.01\\
\quad w/ MedCPT  & 87.32 & 83.54 & 85.39 & 91.36 & 91.36 & 91.36 & 78.17 & 67.99 & 72.73 & 88.64 & 88.64 & 88.64 &84.53\\
\quad w/ Contriever & 85.39 & 83.36 & 84.36 & 92.66 & 92.66 & 92.66 & 76.27 & 68.83 & 72.41 & 91.27 & 91.27 & 91.27  & \textbf{85.18}  \\
\quad w/ BMRetriever & 86.74 & 83.73 & 85.21 & 91.58 & 91.58 & 91.58 & 75.62 & 71.23 & 73.36 & 89.08 & 89.08 & 89.08 & 84.81 \\

\textbf{PMC-Llama-7B}\\
\quad w/ random example & 82.46 & 83.03 & 82.74 & 88.05 & 88.05 & 88.05 & 69.76 & 69.27 & 69.51 & 86.03 & 86.03 & 86.03 & 81.58 \\
\quad w/ MedCPT  & 84.74 & 86.51 & 85.62 & 92.44 & 92.44 & 92.44 & 70.94 & 67.39 & 69.12 & 90.83 & 90.83 & 90.83 & 84.50\\
\quad w/ Contriever & 83.45 & 83.45 & 83.45 & 93.09 & 93.09 & 93.09 & 72.58 & 72.07 & 72.32 & 92.58 & 92.58 & 92.58  & \textbf{85.36}  \\
\quad w/ BMRetriever & 84.81 & 85.40 & 85.10 & 91.74 & 91.74 & 91.74 & 72.00 & 72.51 & 72.26 & 89.52 & 89.52 & 89.52 & 84.66\\
\bottomrule
\end{tabular}
}
\vspace{+2mm}
\caption{Performance of RAMIE. The highest F1 score for each task and the best average F1 score for each model are presented in bold.} 
\label{tab:rmtl}
\end{table*}

\section{Results}
\subsection{Instruction Fine-tuning Performance}
Table \ref{tab:stlzt} presents a comprehensive comparison of various models evaluated under single-task instruction-tuning setting across four tasks.
Overall, compared to BERT, all the LLMs achieved F1 scores above 80, significantly surpassing BERT’s average score of 69.55. This highlights the superior performance of LLMs on DS-related tasks. The key difference lies in the NER task: while BERT classifies each token individually, LLMs extract information directly from entire sentences, leading to a substantial performance gap.
For LLMs, Table \ref{tab:stlzt} shows that the MedAlpaca-7B model achieves the highest average F1-score of 86.38, outperforming other models such as BioMistral-7B (85.36) and Mistral-7B (85.06). This indicates that MedAlpaca-7B is particularly effective when fine-tuned for individual tasks. 

As for the individual task, for example, MedAlpaca-7B achieved an F1 score of 88.57 in NER task. Llama3-8B demonstrated exceptional performance in the RE task, with an F1 score of 93.61. In the TE task, MedAlpaca-7B scored 75.36, while BERT achieved the highest F1 score of 92.61 in the UC task, with LLM models performing closely behind.
Notably, traditional models like BERT lag significantly behind, highlighting the advancements brought by more recent architectures.

\subsection{RAMIE Performance}
Table. \ref{tab:rmtl} presents the performance of 8 LLMs across NER, RE, TE, and UC. 
Each row reflects the performance of a single model for each dataset.
Across all models, the BMRetriever configuration consistently yields the highest average F1 scores. For example, BioMistral-7B achieves its best average F1 score of 85.29 when using BMRetriever. Similarly, Llama2-7B, MedAlpaca-7B, and Mistral-7B also show their best performance under the BMRetriever setting. Notably, Llama2-13B achieves its highest score of 85.96 using Contriever, indicating a strong performance in this configuration.

In terms of performance of individual tasks, RE generally demonstrates higher scores across the models compared to the other tasks, with several models achieving F1 scores above 90. For instance, Llama2-13B under BMRetriever achieves an F1 score of 93.74 in RE. TE however tends to have more variable results, with F1 scores ranging from around 60 to 73 across most models, suggesting the relative difficulty of this task.


\subsection{Ablation Study} 
Table \ref{tab:mtlzt} summarizes the evaluation results under the multi-task instruction fine-tuning setting without RAG. 
Among the models, Mistral-7B stands out with the highest average F1 score of 84.06, surpassing BioMistral-7B (82.10) and Llama3-8B (83.18), indicating its strong and consistent performance across tasks. Most models perform reliably in the RE task, with several exceeding an F1 score of 90. Notably, Llama3-8B, BioMistral-7B, and MedAlpaca-7B all achieved F1 scores of 92.22 or higher in this task. However, the TE task exhibits greater variability in F1 scores, with Mistral-7B achieving the highest at 70.74, highlighting the relative difficulty of this task.

Additionally, we computed the relative F1 score drop for each task and model, comparing to single-task instruction fine-tuning. Under the multi-task setting, the MedAlpaca-13B model experienced the largest drop of 5.66\%, while the average performance drop across all models was 2.72\%. Specifically, for the NER, TE, and UC tasks, the average drops across the eight LLMs were 2.77\%, 5.64\%, and 3.31\%, respectively. Interestingly, the RE task showed a slight benefit from the multi-task instruction fine-tuning.

\begin{table*}[!h]
\centering
\renewcommand\arraystretch{1.3}
\resizebox{1\textwidth}{!}{%
\begin{tabular} {lcccccccccc}
\toprule 
\multicolumn{1}{c}  {}&\multicolumn{2}{c}  {NER} &\multicolumn{2}{c}  {RE} &\multicolumn{2}{c}  {TE} & \multicolumn{2}{c}  {UC} & \multicolumn{2}{c}  {Average}\\
\cmidrule(lr){2-3} \cmidrule(lr){4-5} \cmidrule(lr){6-7} \cmidrule(lr){8-9} \cmidrule(lr){10-11}
 \multicolumn{1}{c} {Models} &  F1 & Perf. Drop &  F1 & Perf. Drop &  F1 & Perf. Drop &  F1 & Perf. Drop & F1 & Perf. Drop \\ 
\hline
\textbf{BioMistral-7B} & 81.52 &	5.15\% &	92.22 &	0.93\% &	68.61 &	4.16\% &	86.03 	& 5.24\% &	82.10 	&3.82\% \\
\textbf{Llama2-7B} & 81.07 &	2.98\% &	87.90 &	-5.43\% &	62.54 &	8.19\% &	88.21 	& 1.95\% &	79.93 &	1.62\% \\
\textbf{Llama2-13B} & 83.30 &	1.21\% &	92.22 &	0.47\% &	67.12 &	5.85\% &	86.03 	& 6.18\% &	82.17 &	3.32\% \\
\textbf{MedAlpaca-7B} & 81.95 &	3.89\% &	90.28 &	3.56\% &	70.66 &	4.68\% 	& 89.08 &	2.12\% &	82.99 &	3.51\% \\
\textbf{MedAlpaca-13B} & 82.10 	& 7.30\% &	90.06 &	-1.16\% &	66.03 &	12.38\% &	87.77 	& 5.19\% &	81.49 &	5.66\% \\
\textbf{Llama3-8B} & 83.85 &	0.11\% &	92.22 &	-3.13\% &	69.32 &	3.29\% &	87.33 	& 3.85\% &	83.18 &	0.94\% \\
\textbf{Mistral-7B} & 83.32 &	1.30\% &	91.79 &	0.94\% &	70.74 &	4.52\% &	90.39 &	-1.47\% &	84.06 &	1.18\% \\
\textbf{PMC-Llama-7B} &  83.95 &	0.19\% &	89.42 &	1.19\% &	65.60 &	2.05\% &	86.03 &	3.42\% &	81.25 &	1.72\% \\
\bottomrule
\end{tabular}}
\caption{Performance of multi-task instruction fine-tuning setting without RAG. The performance drop is compared with single-task instruction fine-tuning.}
\label{tab:mtlzt}
\end{table*}

\begin{figure*}[t]
    \centering
\includegraphics[width=0.99\linewidth]{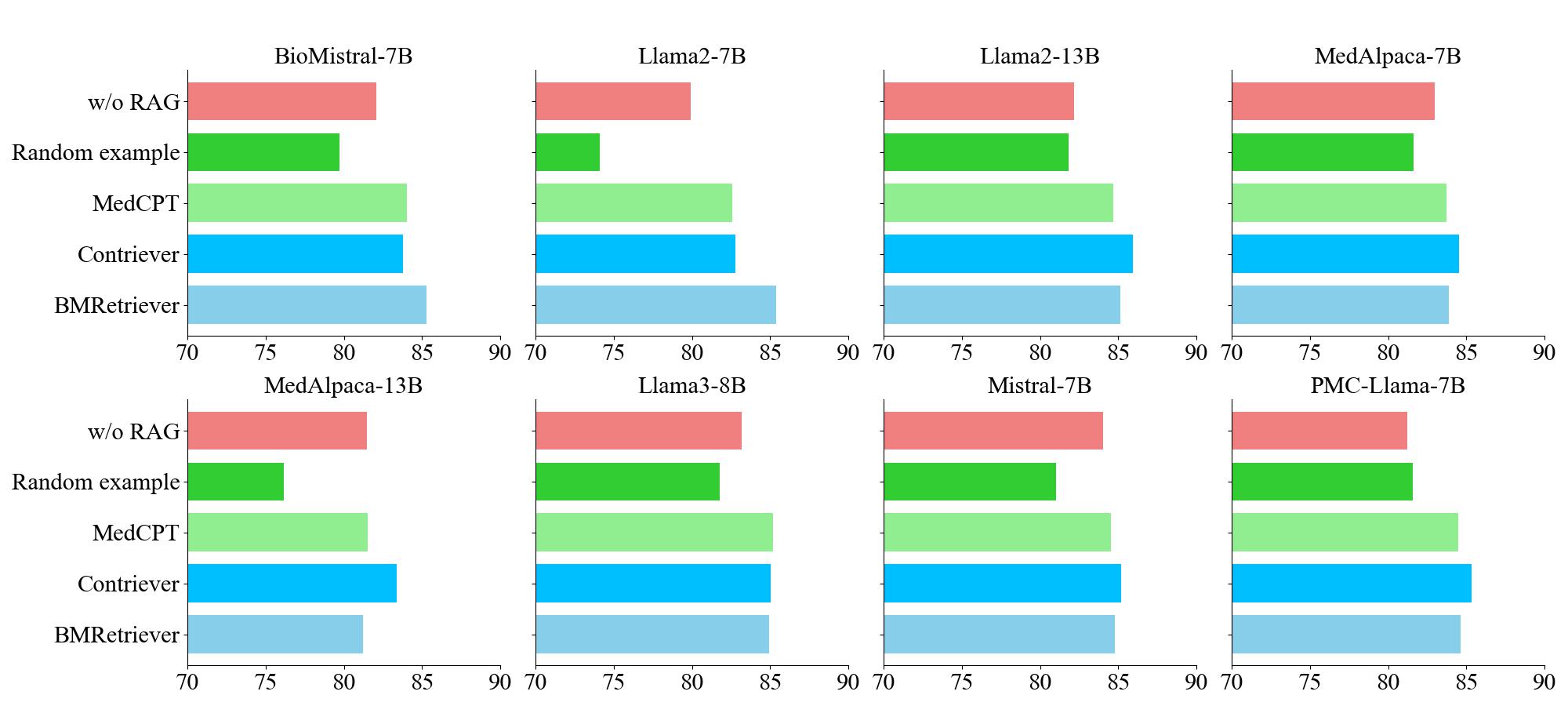}
    \caption{Performance comparison (average F1 score) of RAMIE framework and the MTL without RAG. We averaged the F1 scores of four tasks for each model.}
    \label{fig:mtlrag}
\end{figure*}

\section{Discussion}
Extracting DS information in clinical notes is crucial due to the widespread and growing use of DSs and the need for accurate information on their usage, efficacy, and safety. Given limited regulatory oversight, extracting reliable DS-related information from clinical records is essential for patient safety and informed decision-making. LLMs, with their advanced NLP capabilities, are uniquely suited to this task. Their ability to manage complex, unstructured data allows them to efficiently handle multiple DS-related tasks. By leveraging techniques like instruction fine-tuning, MTL, and RAG, LLMs enhance the precision and relevance of extracted DS data, making them a transformative tool for advancing DS safety and effectiveness in healthcare. In this section, we reflect on the results in the context of single-task instruction fine-tuning, the RAMIE framework, and the effects of removing RAG from the framework. In the end, we analyzed errors and discussed the limitations and future directions.

\begin{figure*}[t]
    \centering
    \includegraphics[width=0.8\linewidth]{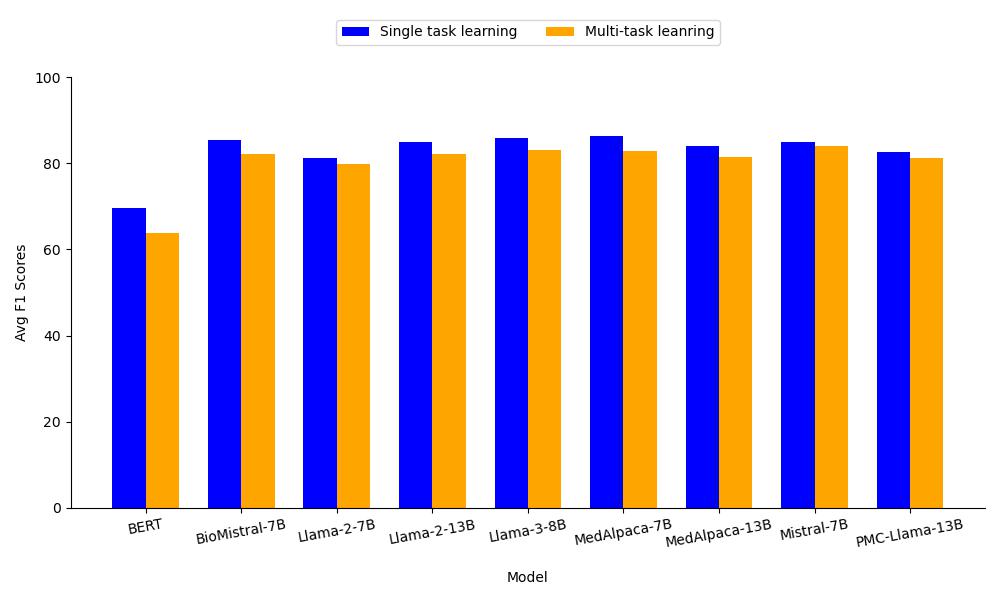}
    \caption{Average F1 scores comparison between single-task learning and multi-task learning.}
    \label{fig:stlvsmtl}
\end{figure*}

The first key finding from this study is that LLMs demonstrated exceptional performance on DS-related tasks, significantly outperforming BERT-based models. For example, MedAlpaca-7B achieved the highest F1 score (86.38) under the single-task instruction fine-tuning setting, which shows the superiority of LLMs in handling complex, multi-faceted tasks. 
Compared to BERT, LLMs have a more advanced architecture and larger model size, allowing them to capture deeper semantic patterns and make use of richer contextual information. Moreover, LLMs’ generative nature enables better adaptability across diverse tasks without requiring specific architecture changes, which is a limitation in smaller models like BERT. The ability of LLMs to generalize across tasks while maintaining high performance makes them particularly well-suited for complex domains such as biomedical information extraction.

Second, our RAMIE provided a framework for MTL, achieving comparable performances of single-task instruction fine-tuning. For instance, Llama2-13B, when combined with the Contriever retriever, reached an F1 score of 85.96, outperforming the single-task fine-tuning counterpart (F1 of 84.99).
We conducted a comprehensive analysis of all experiments, comparing the performance of models in single-task settings and models with RAG across four tasks. This included combinations of each model with three different retrievers, totaling 96 experimental setups. Among these, 46 setups achieved better results using our framework, demonstrating that the outcomes from our framework are comparable to those from single-task instruction fine-tuned models.
The key advantage of the RAMIE framework lies in its use of RAG, which enhances LLMs by providing additional examples during training and testing. RAG helps the model retrieve relevant examples that augment the training data, thus enabling more accurate output generation. 
As shown in Fig. \ref{fig:mtlrag}, we compared the performance of each LLM across five settings: zero-shot (labeled as 'w/o RAG' in table), random example, and three retrievers (MedCPT, Contriever, and BMRetriever). Zero-shot serves as the baseline without RAG and the random example setting serves as the baseline for using examples. 
Interestingly, using a random example in the prompt actually decreases model performance, while all retrievers—MedCPT, Contriever, and BMRetriever—consistently improved the performance of LLMs. 
Contriever shows the best overall performance, with six out of eight models achieving their highest scores when combined with Contriever. BMRetriever follows closely, yielding the best results in two models.
By dynamically fetching task-specific examples, RAG alleviates the burden on the LLM to store all knowledge internally, allowing the model to focus on generating more contextually accurate predictions in DS-related information extraction.

In addition, except that the performances of single-task instruction fine-tuning and our RAMIE framework are comparable, the model trained in our framework demonstrates excellent storage efficiency and reduction on training cost, since a single model is trained once and then can handle all four tasks simultaneously, which verifies the effectiveness of the proposed RAMIE framework.
By reducing the storage requirements and training cost while maintaining or even improving performance for some models, this approach is particularly suited for resource-constrained environments where running LLMs on multiple devices is not feasible.

Lastly, we conducted an ablation study where we ran MTL experiments (without RAG). When comparing the results to single-task instruction fine-tuning, we observed that the overall average F1 scores decreased, as shown in Fig. \ref{fig:stlvsmtl}.
The reason behind this decline is that MTL requires the model to manage multiple tasks simultaneously, which leads to increased complexity in distinguishing between tasks and adjusting to varying output formats. This cognitive load on the model can cause a negative influence, where learning multiple tasks hinders the model's performance on individual tasks. The need to allocate capacity to learn the differences between tasks, as well as the varying output formats for each task, leads to a slight drop in performance. This is consistent with findings from previous research~\cite{mulyar2021mt}.

\subsection*{Error analysis}
Diving into these promising results, through observing the generation, we found LLMs are suffering from the following errors:
\begin{itemize}
\item Redundant Information: In tasks like NER and TE, the language model occasionally extracts unnecessary information. For instance, while expert annotations label "motion sickness" as an adverse event, the model often extracts "mild motion sickness" as the result. Here, "mild" is an adjective describing the adverse event rather than a core part of it.

\item Information omission: The model occasionally failed to capture all relevant information. In sentences with multiple entities, it might extract only a subset of them—capturing three out of four entities, for instance. Similarly, when there are multiple triples in a sentence, the model sometimes retrieves only one, overlooking additional relevant relationships.

\item Incorrect generation: It includes misclassifying entities, incorrectly assigning relationships between head and tail entities, and extracting non-entity words as entities.
\end{itemize}


\subsection*{Limitations and future Directions}
This work has a few limitations. First, we did not include particularly large models (e.g., Llama-2-70B) in our experiments. Larger models often have greater capacity, which may enable them to better manage the complexities of learning multiple tasks concurrently, potentially enhancing MTL performance. Second, to our knowledge, no retriever exists that is specifically optimized for DS tasks, which might further improve model accuracy and relevance in this domain. Lastly, we did not investigate few-shot learning scenarios. Examining the model’s performance under few-shot conditions could offer valuable insights into its adaptability and ability to generalize in contexts with limited labeled data.
For future research, we recommend further exploration of retrieval-augmented multi-task frameworks using various LLM architectures and datasets. This could provide a deeper understanding of the framework's scalability across different domains. Additionally, experimenting with a broader range of retrievers beyond MedCPT, Contriever, and BMRetriever may uncover more effective methods to enhance model performance. Finally, investigating transfer learning between related tasks or integrating more advanced model architectures could help address the observed performance limitations in MTL, leading to improved efficiency and accuracy across tasks.

\section{Conclusion}
This paper presented the RAMIE framework, a retrieval-augmented, multi-task LLM solution for extracting dietary supplement information from clinical records. RAMIE effectively handled NER, RE, TE, and UC tasks, achieving high accuracy and efficiency through instruction fine-tuning, MTL, and RAG. Experimental results demonstrated that RAMIE demonstrates its excellent abilities in DS information extraction while achieving efficiency, confirming its potential for scalable, efficient information extraction in healthcare applications.

\section{Contributorship Statement}
Zaifu Zhan and Rui Zhang conceptualized and designed the study. 
Zaifu Zhan and MingChen Li curated the data.
Zaifu Zhan executed the experiments. 
Zaifu Zhan and Shuang Zhou drafted the initial manuscript, and Rui Zhang reviewed and finalized the manuscript.
Rui Zhang supervised the whole project.

\section{Funding Statement}
This work was supported by the National Institutes of Health’s National Center for Complementary and Integrative Health under grant numbers R01AT009457 and U01AT012871, the National Institute on Aging under grant number R01AG078154, the National Cancer Institute under grant number R01CA287413, and the National Institute of Diabetes and Digestive and Kidney Diseases under grant number R01DK115629. The content is solely the responsibility of the authors and does not represent the official views of the National Institutes of Health.

\section{Acknowledgements}
We would like to express our sincere gratitude to the reviewers for any suggestions.

\section{Competing Interests Statement}
The authors state that they have no competing interests to declare.












\bibliography{0_main}
\bibliographystyle{iclr2021_conference}


\end{document}